# TURN SEGMENTATION INTO UTTERANCES FOR ARABIC SPONTANEOUS DIALOGUES AND INSTANT MESSAGES

AbdelRahim A. Elmadany[1], Sherif M. Abdou[2] and Mervat Gheith[1]

[1] Institute of Statistical Studies and Research (ISSR), Cairo University
[2] Faculty of Computers and Information, Cairo University

## ABSTRACT

*Text segmentation task is an essential processing task for many of Natural Language Processing (NLP) such as text summarization, text translation, dialogue language understanding, among others. Turns segmentation considered the key player in dialogue understanding task for building automatic Human-Computer systems. In this paper, we introduce a novel approach to turn segmentation into utterances for Egyptian spontaneous dialogues and Instance Messages (IM) using Machine Learning (ML) approach as a part of automatic understanding Egyptian spontaneous dialogues and IM task. Due to the lack of Egyptian dialect dialogue corpus the system evaluated by our corpus includes 3001 turns, which are collected, segmented, and annotated manually from Egyptian call-centers. The system achieves $F_1$ scores of 90.74% and accuracy of 95.98%.*

## 1.INTRODUCTION

Build a completely Human-Computer systems and the belief that will happens has long been a favourite subject in research science. So, dialogue language understanding is growing and considering the important issues today for facilitating the process of dialogue acts classification; consequently segment the long dialogue *turn* into meaningful units namely *utterances* are increasing.

This paper refers to an *utterance* as a small unit of speech that corresponds to a single act[1,2]. In speech research community, *utterance* definition is a slightly different; it refers to a complete unit of speech bounded by the speaker's silence while, we refer to the complete unit of speech as *a turn*. Thus, a single *turn* can be composed of many *utterances*. *Turn* and *utterance* can be the same definition when the *turn* contains one utterance as defined and used in [3] .

Our main motivation for the work reported here comes from automatic understanding Egyptian dialogues and IM which called "dialogue acts classification". Dialogue Acts (DA) are labels attached to dialogue utterances to serve briefly characterize a speaker's intention in producing a particular utterance [1].

           

Egyptian *turns* are almost long and contains many *utterances* as we noticed during data collection. Consequently, we propose a novel approach to *turn* segmentation into *utterances* for Egyptian Arabic and Arabic Instant Messages (IM) namely 'USeg', which has not addressed before to the best of our knowledge.

USeg is a machine learning approach based on context without relying on punctuation, text diacritization or lexical cues. Whereas, USeg depends on a set of features from the annotated data that's include morphological features which have been determined by the Morphological Analysis and Disambiguation of Arabic Tool (MADAMIRA)[1] [4]. USeg is evaluated by an Arabic dialogue corpus contains spoken dialogues and instant messages for Egyptian Arabic, and results are compared with manually segmented turns elaborated by experts.

This paper is organized as follows: section 2 present the Egyptian dialect, section 3 present the background, section 4 describe the corpus used to experiment, section 5 present the proposed approach "USeg", section 6 present the experimental setup and results; and finally the conclusion and feature works is reported in section 7.

## 2. ARABIC LANGUAGE

Arabic is one of the six official languages of the United Nations. According to Egyptian Demographic Center, it is the mother tongue of about 300 million people (22 countries). There are about 135.6 million Arabic internet users until 2013[2].

The orientation of writing is from right to left and the Arabic alphabet consists of 28 letters. The Arabic alphabet can be extended to ninety elements by writing additional shapes, marks, and vowels. Most Arabic words are morphologically derived from a list of roots that are tri, quad, or pent-literal. Most of these roots are tri-literal. Arabic words are classified into three main parts of speech, namely nouns, including adjectives and adverbs, verbs, and particles. In formal writing, Arabic sentences are often delimited by commas and periods. Arabic language has two main forms: Standard Arabic and Dialectal Arabic. Standard Arabic includes Classical Arabic (CA) and MSA while Dialectal Arabic includes all forms of currently spoken Arabic in daily life, including online social interaction and it vary among countries and deviate from the Standard Arabic to some extent[5]. There are six dominant dialects, namely; Egyptian, Moroccan, Levantine, Iraqi, Gulf, and Yemeni.

MSA considered as the standard that commonly used in books, newspapers, news broadcast, formal speeches, movies subtitles,… etc.. Egyptian dialects commonly known as Egyptian colloquial language is the most widely understood Arabic dialects due to a thriving Egyptian television and movie industry, and Egypt's highly influential role in the region for much of the 20th century[6]. Egyptian dialect has several large regional varieties such as Delta and Upper Egypt, but the standard Egyptian Arabic is based on the dialect of the Egyptian capital which is the most understood by all Egyptians.

---

[1] http://nlp.ldeo.columbia.edu/madamira/
[2] http://www.internetworldstats.com/stats7.htm

## 3. BACKGROUND

A segmentation process generally means dividing the long unit, namely "*turn*" into meaningful pieces or small units "non-overlapping units" namely "*utterances*". Moreover, we distinguish three main approaches to turn segmentations:

− The acoustic segmentation approach is usually segmented the long input "waveform" into short pieces based on acoustic criteria features such as pauses "non-speech intervals".

− Linguistic segmentation is segment the turn based on syntactic and semantic features such as morphological features.

− The mixed approach is used the acoustic and linguistic features.

Due to the lack of an Egyptian Arabic recognition system, manual transcription of the corpus is then required. Therefore, we focus on linguistic segmentation for Arabic spontaneous dialogues and an IM segmentation task that has several challenges:

− Essential characteristics of spontaneous speech: ellipses, anaphora, hesitations, repetitions, repairs… etc. These are some examples from our corpus:

  o  A user who does repairs and apologize in his turn: السفر يوم 12 ديسمبر اسفه 11 ديسمبر (*Alsfr ywm 12 dysmbr Asfh 11 dysmbr*, the arrival on 12 sorry 11 December)[3].
  o  A user who repeats the negative answer and produce non-necessary information on his turn: لا لا انا مش فاتحة حساب عندكم وانا مبشتغلس بس جوزي هو اللي بيشتغل ( *lA lA AnA m$ fAtHp HsAb Endkm wAnA mb$tgls bs jwzy hw Ally by$tgl*, No No I don't have an account in your bank and I'm not an employee but my husband is an employee)

− **Code Switching:** using a dialect words which are derived from foreign languages by code switching between Arabic and other language such as English, France, or Germany. Here an example for user who uses foreign "Egnlish" words in his turn such as ترانزاكشن (*trAnzAk$n*, Transaction) and اكتف (*Aktf*, Active) in his turn. اممم فده متاح ولا لازم من الاول اعمل اي ترانزاكشن علشان يبقي اكتف بدل دورمنت (*Ammm fdh mtAH wlA lAzm mn AlAwl AEml Ay trAnzAk$n El$An ybqy Aktf bdl dwrmnt*, Um this is available or I need to do any transaction to activate the dormant account)

− **Deviation:** Dialect Arabic words may be having some deviation such as MSA "اريد" (*Aryd*, want) can be "عايز" (*EAyz*, want), or "عاوز" (*EAwz*, want) in Egyptian dialect.

− **Ambiguity:** Arabic word may be having different means such as the word "علم" can be: "عَلَم" "*flag*", "عِلْم" "*science*", "عُلِمَ" "*it was known*", "عَلِمَ" "*he knew*", "عَلَّمَ" "*he taught*" or "عُلِّمَ" "*he was taught*". Thus, the ambiguity considers the key problem for Natural

---

[3] Examples are written as Arabic (*Buckwalter transliteration schema*, English translation)

International Journal on Natural Language Computing (IJNLC) Vol. 4, No.2,April 2015

Language Understanding / Processing especially on the Arabic language. The word diacritization is very useful to clarify the meaning of words and disambiguate any vague spellings.

− **Lack of Resources:** The not existence and the lack of tagged Arabic Spontaneous Dialogues and Instance Messages corpora for Egyptian Arabic corpus make turn segmentation task far more challenging. Since manual construction of tagged corpus is time-consuming and expensive [25], it is difficult to build large tagged corpus for Arabic dialogues acts and turn segmentation. So, the researchers had to build their own resources for testing their approaches. Consequently, we built our corpus and used it for both training and testing (see section 4).

The most of turn segmentation into utterances approaches such as[7-10] are developed and tested on non-Arabic languages such as English, Germany or France. There are few works interested in Arabic dialogue acts classification; these works have defined and used the user's *turn* as an *utterance* without any segmentation such as [3,11-15]. However, there are some approaches used for Arabic text segmentation based on linguistic approaches either rule-based such as [16-18] or machine learning based such as [19]. These approaches mainly rely on punctuation indicators/marks, conjunctions, text diacritization or/and lexical cues. In addition, it is designed and applied on MSA text such as newspapers and books, which are completely different from Arabic spontaneous dialogues and IM, which are considered an informal [20].

## 4. CORPUS CONSTRUCTION

We built our own corpus namely JANA, which has manually dialogues turns segmented into utterances and annotated with dialogue acts schema. JANA is a multi-genre corpus of Arabic dialogues labeled for Arabic Dialogues Language Understanding (ADLU) at utterance level ant it comprises spontaneous dialogues and IM for Egyptian dialect. Building JANA corpus proceeds in three stages:

1. In the first stage, we collected/recorded 200 dialogues manually from different genre call centers such as Banks, Flights, Mobile Network Providers (MNP), and MNP's online-support using Egyptian native speakers since August 2013; these dialogues consist of human-human conversation and instant messages about inquiries regarding providing service from call centers such as create a new bank account, service request, balance check and flight reservation.

2. In the second stage, we are randomly choice 52 spoken dialogues from recorded dialogues and 30 IM dialogues as the first release of JANA corpus. The selected phone calls of spoken dialogues are recorded manually with an average duration of two hours of talking time after removing ads and waiting times from phone calls. Moreover, these phone calls are transcribed using Transcriber[4] toolkit.

3. In the third stage, turns are segmented into utterances manually and the utterances are tagged with DAs labels which reported by [21] manually.

---

[4] Available at http://trans.sourceforge.net/en/presentation.php

114

International Journal on Natural Language Computing (IJNLC) Vol. 4, No.2,April 2015

Table 1. Main characteristics of the first release of JANA corpus

| Total Number of Dialogues | 82 |
|---|---|
| Spoken Dialogues Number | |
|     \|------ Banks | 26 |
|     \|------ Flights | 26 |
| Written Dialogues (Chats) Number | |
|     \|------ Mobile Network Operators | 30 |
| Total Number of Turns | 3,001 |
| Number of Segmented Turns | 1,091 |
| Number of Utterances from Segmented Turns | 2,815 |
| Total Number of Utterances | 4,725 |
| Words | 20,113 |
| Words per Turn | 6.7 |
| Words per Utterance | 4.3 |

The first release of JANA consists of approximately 3001 turns with average 6.7 words per turn, contains 4725 utterances with average 4.3 words per utterance, and 20311 words; and it will be made freely available to academic and nonprofit research. Moreover, the most important characteristics of JANA corpus are shown in Table 1 and a sample of turn's segmentation and utterance's DAs annotation is shown in Table 2. In addition, dialogue sample before segmentation and DAs annotation process is shown in Table 3.

Table2. Fully turns segmented and DAs tagged sample from JANA corpus

| Turn ID | Persons | Utterance ID | Utterances | | Dialogue Act |
|---|---|---|---|---|---|
| T1 | Operator | U1 | Arabic:<br>Buckwalter:<br>English: | ان اس جي بي<br>*An As jy by*<br>NSBG | SelfIntroduce |
| | | U2 | Arabic:<br>Buckwalter:<br>English: | شريفة المصري<br>*$ryfp AlmSry*<br>Sherifa Elmasri | SelfIntroduce |
| | | U3 | Arabic:<br>Buckwalter:<br>English: | مساء الخير<br>*msA' Alxyr*<br>Good afternoon | Greeting |
| T2 | Customer | U4 | Arabic:<br>Buckwalter:<br>English: | الو<br>*Alw*<br>Allo | Taking_Request |
| | | U5 | Arabic:<br>Buckwalter:<br>English: | مساء الخير<br>*msA' Alxyr*<br>Good afternoon | Greeting |
| T3 | Operator | U6 | Arabic:<br>Buckwalter:<br>English: | مساء النور<br>*msA' Alnwr*<br>Good afternoon | Greeting |
| T4 | Customer | U7 | Arabic:<br>Buckwalter:<br>English: | من فضلك<br>*mn fDlk*<br>If you please | Taking_Request |
| | | U8 | Arabic:<br>Buckwalter:<br>English: | كنت عايزة اسأل عن قروض السيارات<br>*knt EAyzp As>l En qrwD AlsyArAt*<br>I want to ask about cars loan | Service_Question |
| | | U9 | Arabic:<br>Buckwalter:<br>English: | بس هو المشكلة انني معنديش اصلا حساب عندكم<br>*bs hw Alm$klp Anny mEndy$ ASlA HsAb Endkm*<br>The problem is I haven't an account in your bank | Inform |
| ⋮ | ⋮ | ⋮ | | ⋮ | ⋮ |
| T13 | Operator | U20 | Arabic:<br>Buckwalter:<br>English: | اي استفسار تاني<br>*Ay AstfsAr tAny*<br>Any other service? | Confirm_Question |
| T14 | Customer | U21 | Arabic:<br>Buckwalter:<br>English: | ميرسي<br>*myrsy*<br>No thanks | Disagree |
| T15 | Operator | U22 | Arabic:<br>Buckwalter:<br>English: | شكرا علي اتصال حضرتك<br>*$krA Ely AtSAl HDrtk*<br>Thanks for your calling | Greeting |

Table 3. Sample of dialogue before segmentation and DAs annotation process

| Turn ID | Persons | | Turns | |
|---|---|---|---|---|
| T1 | Operator | Arabic:<br>*Buckwalter:*<br>English: | | مساء الخير بنك مصر احمد مع حضرتك<br>*msA' Alxyr bnk mSr AHmd mE HDrtk*<br>Good evening, Banque Misr, Ahmed speaking |
| T2 | Customer | Arabic:<br>*Buckwalter:*<br>English: | | السلام عليكم<br>*AslAm Elykm*<br>Hello |
| T3 | Operator | Arabic:<br>*Buckwalter:*<br>English: | | عليكم السلام<br>*Elykm AslAm*<br>Hello |
| T4 | Customer | Arabic:<br>*Buckwalter:*<br>English: | | معاك محمد صفوت علي<br>*mEak mHmd Sfwt Ely*<br>Mohamed Safwat Ali speaking |
| T5 | Operator | Arabic:<br>*Buckwalter:*<br>English: | | تمام اهلا يا استاذ محمد<br>*tmAm AhlA yA AstA* mHmd*<br>Ok, welcome Mr. Mohamed |
| T6 | Customer | Arabic:<br>*Buckwalter:*<br>English: | | كنت عاوز اسأل حضرتك عن خطوات الاشتراك في خدمت الانترنت البنكي<br>*knt EAwz As>l HDrtk En xTwAt AlA$trAk fy xdmt AlAntrnt Albnky*<br>I want to ask about the steps to participate in online banking |

| | | | |
|---|---|---|---|
| T7 | Operator | Arabic:<br><br>Buckwalter:<br><br>English: | خطوات الانترنت البنكي استاذ محمد حضرتك تشرفنا في الفرع اقرب فرع لحضرتك وبتقدم طلب للاشتراك في خدمة الاونلاين تملأ البيانات بتاعة حضرتك وخلال اسبوعين عمل بيتم اصدار يوزر اي دي<br>xTwAt AlAntrnt Albnky AstA* mHmd HDrtk t$rfnA fy AlfrE Aqrb frE lHDrtk wbtqdm Tlb llA$trAk fy xdmp AlAwnlAyn tmlA AlbyAnAt btAEp HDrtk wxlAl AsbwEyn Eml bytm ASdAr ywzr Ay dy<br>Online banking steps Mr. Mohamed, you can go to nearest branch of your presence and you fill the request to participate in the online service. After two weeks, we will send to you the username and the password. |
| T8 | Customer | Arabic:<br>Buckwalter:<br>English: | في السيت بتاع بنك مصر<br>fy Alsyt btAE bnk mSr<br>In the bank website. |
| T9 | Operator | Arabic:<br>Buckwalter:<br>English: | في اقرب فرع لحضرتك<br>fy Aqrb frE lHDrtk<br>No, In nearest branch |
| T10 | Customer | Arabic:<br>Buckwalter:<br>English: | تمام اه مش ممكن من خلال النت يعني لازم من خلال اقرب فرع<br>tmAm Ah m$ mmkn mn xlAl Alnt yEny lAzm mn xlAl Aqrb frE<br>Uh, there is not possible to do it through the net or that is necessary through the nearest branch |
| T11 | Operator | Arabic:<br>Buckwalter:<br>English: | اه لازم تقدم من الفرع<br>Ah lAzm tqdm mn AlfrE<br>Through the nearest branch |
| T12 | Customer | Arabic:<br>Buckwalter:<br>English: | تمام<br>tmAm<br>OK |
| T13 | Operator | Arabic:<br><br>Buckwalter:<br><br>English: | خلال اسبوعين عمل بيتم اصدار اليوزر وتقدر تستخدمها من خلال خدمة الاونلاين عادي<br>xlAl AsbwEyn Eml bytm ASdAr Alywzr wtqdr tstxdmhA mn xlAl xdmp AlAwnlAyn EAdy<br>It will takes two weeks and you will get the username. |
| T14 | Customer | Arabic:<br>Buckwalter:<br>English: | يبقي فيه يوزرنيم وباسورد وكده صح<br>ybbqy fyh ywzrnym w bAswrd wkdh SH<br>I'll take username and password, right? |
| T15 | Operator | Arabic:<br>Buckwalter:<br>English: | الباسورد بتكريتها علي الموقع<br>AlbAswrd btkrythA Ely AlmwqE<br>No, username only and the password you will create it through the website |
| T16 | Customer | Arabic:<br>Buckwalter:<br>English: | اه تمام كده ماشي شكرا لحضرتك<br>Ah tmAm kdh mA$y $krA lHDrtk<br>Ok, than you |
| T17 | Operator | Arabic:<br>Buckwalter:<br>English: | اي استفسار تاني<br>Ay AstfsAr tAny<br>Any other service can I do for you. |
| T18 | Customer | Arabic:<br>Buckwalter:<br>English: | لا شكرا<br>lA $krA<br>No thanks |
| T19 | Operator | Arabic:<br>Buckwalter:<br>English: | شكرا لسيادتك والسلام عليكم<br>$krA lsyAdtk wAlslAm Elykm<br>Thank you for calling and goodbye |

## 5. METHODOLOGY

Support Vector Machine (SVM) is a supervised machine learning that has been shown to perform well on text classification tasks, where data is represented in a high dimensional space using sparse feature vectors [22,23]. Moreover, the SVM is robust to noise and the ability to deal with a large number of features effectively [24]. The SVM classifier is trained to discriminate between

International Journal on Natural Language Computing (IJNLC) Vol. 4, No.2,April 2015

examples of each class, and those belonging to all other classes combined. During testing, the classifier scores on an example, are combined to predict its class label [25].

USeg is a SVM approach which a Machine learning based involve a selected set of features, extracted from segmented and annotated datasets, which is used to generate a statistical model for segmentation prediction. We used YamCha SVM toolkit[5] that converts the text segmentation task to a text chunking task.

There are three processes to do as preprocessing the input *turns* before running the USeg classifier. These processes are:

1. **Normalization:** to avoid writing errors from the transcription, we normalized the transcribed turns (unified Arabic characters) as

    a. Convert *Hamza-under-Alif* "إ", *Hamza-over-Alif* "أ", and *Madda-over-Alif* "آ" to *Alif* "ا"
    b. Convert *Teh-Marbuta* "ة" to *Heh* "ه"
    c. Convert *Alif-Maksura* "ى" to *Yeh* "ي".

2. Split "و" (*w*, and) from the original words: Sometimes the writers write the conjunction "و" (*w*, and) concatenated to the next word. For instant, "وقال" (***w**qAl*, **and** he talked) the original word "قال" (*qAl*, he talked) is concatenated with the conjunction "و" (*w*, and). To detect and split this "و" (*w*, and) from the original words; we build a tool, namely *Wawanizer*, which is a lookup-table based classifier contains approximately 22K normalized word extracted from news articles and tweets (113,969 Arabic words).

3. The turns are transliterated from Arabic to Latin based ASCII characters using the Buckwalter transliteration scheme[6].

There are two phases has employed for carrying out the classification task in our approach, training phase and test phase. The training phase generates the classification model using a set of classification features. In the test phase, the classification model is utilized to predict a class for each token (word).

In the training phase, each word is represented by a set of features and its actual segmentation state (either "*B-Seg*" to indicate that word is a segment/utterance start or "*I-Seg*" to indicate that word is inside the segment/utterance) in order to produce an SVM model that's able to predict the start of a segment / utterance. Thus, the first step in our approach is to extract the significant features from the training data. Consequently, we study the impact of the features individually by using only one feature at a time and measure the classifier's performance using the F-measure metric. Finally, according to the performance achieved, we select the optimized features for the proposed approach "*USeg*".

---

[5] Available at http://chasen.org/~taku/software/yamcha/
[6] http://languagelog.ldc.upenn.edu/myl/ldc/morph/buckwalter.html

## 5.1. Features Selection

Feature selection refers to the task of identifying a useful subset of features chosen to represent elements of a larger set.

- **Contextual word:** The features of a sliding window, including a word *n*-gram that includes the candidate word, along with previous and following words. For instance, in the training corpus the word "عايز" (*EAyz*, want) appears frequently before a user's request that indicate a request act or new segment/utterance. Therefore, the classifier will use this information to predict a new segment/utterance after this word.

- **Part-Of-Speech (POS):** A value indicating the POS tag is a conjunction, noun, or proper noun such as:

    o **Conjunctions:** For instance, in the training corpus the conjunction "لكن" (lkn, but) appears frequently before a user's request that indicate a new segment/utterance. Also, the conjunction "و" (w, and) is considered as anomaly, that can define a new segment/utterance or not, the classifier handles this problem using sliding window from conjunction along with previous and following words. For instance, " عايز اعرف ازاى افتح حساب و ايه الإجراءات اللازمة لعمل ده" (EAy AErf AzAY AftH HsAb w Ayh Al<jrA'At AllAzmp lEml dh, I want to open an account and what is required steps). In this example conjunction "و" (w, and) separate between two segments/utterances, but here "ايه الفرق بين الحساب الجاري و التوفير" (Ayh Alfrq byn AlHsAb AljAry w Altwfyr, what is the difference between current and saving accounts) the conjunction "و" (w, and) not considered a segment/utterances separator.

    o **Noun or Proper noun:** a binary value defines the word/token is a noun or proper noun. For instance, in the training corpus the service operator introduces himself and his organization comes directly after greeting such as مساء الخير الأهلي فون أحمد سامي (*msA' Alxyr Al>hly fwn >Hmd sAmy*, Good evening Al-Ahly Phone Ahmed Samy ). So, we need to segment this turn into two utterance one includes greeting and the other includes self-introduce.

- **Previous Predicted tags:** The previous words tags of turn can help to can help to anticipate the next the candidate/current word tag. So, we adjust that the SVM predict the tag of the current word using the features and previous predicted tags. The SVM tags can be B-Seg or I-Seg (begin of segment/utterance or inside the segment/utterance). For instance, in the training corpus the word "عايز" (*EAyz*, want) is frequently indicating to new segment/utterance. Sometimes can appears after pronouns that's means the start of the segment/utterance beginning from the pronoun, not from the word "عايز" (*EAyz*, want) such as "انا كنت **عايز** اشترك في الانترنت البنكي" (*AnA knt **EAyz** A$trk fy AlAntrnt Albnky*, I **want** to register for internet banking ).

## 6. EXPERIMENTS AND RESULTS

In order to measure the effect of complexity of each dialogues domain (Banks, Flights, and Mobile Network Operators) on classification accuracy, we experiment on each dialogue domain separately and one experiment to overall combined data. We split each domain based on dialogue turn boundary into 70% training dataset, 20% development dataset (DEV), and the 10 % test dataset as shown in Table 4. The results are reported using standard metrics of Accuracy (*Acc*), Precision (*P*), Recall (*R*), and the $F_1$ score $(F_1)$[7].

Table 4. Corpus training, development (DEV) and test datasets

| | Domain | Datasets | Dialogues | Turns | Utterances |
|---|---|---|---|---|---|
| Spoken | Banks | DEV | 4 | 115 | 193 |
| | | Test | 5 | 226 | 368 |
| | | Training | 17 | 782 | 1,234 |
| | Flights | DEV | 5 | 145 | 242 |
| | | Test | 7 | 224 | 364 |
| | | Training | 14 | 773 | 1,186 |
| IM | Mobile Network Operators | DEV | 3 | 75 | 109 |
| | | Test | 5 | 197 | 272 |
| | | Training | 22 | 464 | 757 |
| | Total | | 82 | 3,001 | 4,725 |

In the training stage, the training is applied on the training dataset using selected features set and the results are analyzed to determine the best features set. The development stage is performed using the DEV dataset to define the best feature set which used in the test stage. In the test stage, the classifier is applied on the test dataset and the results are reported and discussed.

The selected features are tested on window size within ranges from -1/+1 to -5/+5. We found that a context size of -2/+2 (two previous word and two subsequent word) with three of previous predicted tags achieves the best performance in this task.

Table shows the results for each domain (Banks, Flights, and Mobile Networks Operators). These results and the overall experiment shows that USeg classifier yield high performance and efficiency in Arabic dialect dialogue turn segmentation into utterances task.

---

[7] $F_1 = \frac{2PR}{P+R}$

Table 5. Testing Results

|  | Domain | *P* | *R* | *F₁* | *Acc* |
|---|---|---|---|---|---|
| Spoken | Banks | 97.47 | 83.70 | 90.06 | 95.91 |
| Spoken | Flights | 96.38 | 80.50 | 87.72 | 94.44 |
| IM | Mobile Network Operators | 96.57 | 82.72 | 89.11 | 95.47 |
| **Overall Experiment** | | **96.84** | **85.36** | **90.74** | **95.98** |

In this work, we reported some difficulties that we faced.

- The ف (*f*, Fa) and ب (*b*, Ba) conjunctions are considered as the most complex type of conjunctions that cannot detected by any POS tagger such as بحاول (*bHAwl*, I'm trying) consist of ب (*b*, Ba) + حاول (*HAwl*, trying)

- In IM, sometimes the writers write the Arabic word in Franco-Arabic style. For instance, the word "A7med" express the person's name Ahmed and "3ayz" express a dialect word "عايز" (*EAyz*, want).

## 7. CONCLUSION

In this paper, we present a machine-learning approach using SVM to solve the problem of automatic Arabic dialogues turns segmentation into utterances task as a part of Arabic dialogues understanding task for Egyptian dialect; namely, USeg. In addition, we present JANA corpus that a multi-genre corpus of Arabic dialogues labeled in Arabic Dialogues Language Understanding at utterance level ant it is comprised spontaneous dialogues and IM for Egyptian dialect.

The results obtained that USeg classifier is very promising. To the best of our knowledge, these are the first results reported for turn segmentation into utterances task for Egyptian dialect.
We are currently trying to generalize USeg by applying it in news text domain and social networks text domain such as twitter.

# REFERENCE

[1] N. Webb, "Cue-Based Dialogue Act Classification," Ph.D. dissertation, Department of Computer Science, University of Sheffield, England, 2010.
[2] D. Traum and P. A. Heeman, "Utterance units in spoken dialogue," in Dialogue processing in spoken language systems, ed. Berlin Heidelberg: Springer, 1997, pp. 125-140.
[3] M. Graja, M. Jaoua, and L. H. Belguith, "Discriminative Framework for Spoken Tunisian Dialect Understanding," presented at the 2nd International Conference on Statistical Language and Speech Processing, SLSP 2014, 2013.
[4] A. Pasha, M. Al-Badrashiny, M. Diab, A. E. Kholy, R. Eskander, N. Habash, M. Pooleery, O. Rambow, and R. M. Roth, "MADAMIRA: A Fast, Comprehensive Tool for Morphological Analysis and Disambiguation of Arabic," presented at the Language Resources and Evaluation Conference (LREC 2014), 2014.
[5] M. Elmahdy, G. Rainer, M. Wolfgang, and A. Slim, "Survey on common Arabic language forms from a speech recognition point of view," In Proceeding of International conference on Acoustics (NAG-DAGA), 2009, pp. 63-66.
[6] O. F. Zaidan and C. Callison-Burch, "Arabic dialect identification," Computational Linguistics, vol. 52 (1), 2012.
[7] J. Ang, Y. Liu, and E. Shriberg, "Automatic Dialog Act Segmentation and Classification in Multiparty Meetings," In Proceeding of IEEE International Conference on Acoustics, Speech, and Signal Processing (ICASSP '05), 2005, pp. 1061 - 1064.
[8] E. Ivanovic, "Automatic utterance segmentation in instant messaging dialogue," In Proceeding of The Australasian Language Technology Workshop, 2005, pp. 241-249.
[9] M. Zimmermann, Y. Liu, E. Shriberg, and A. Stolcke, "Toward Joint Segmentation and Classification of Dialog Acts in Multiparty Meetings," In Proceeding of Proc. Multimodal Interaction and Related Machine Learning Algorithms Workshop (MLMI–05), 2005.
[10] L. Ding and C. Zong, "Utterance segmentation using combined approach based on bi-directional N-gram and maximum entropy," Proceedings of the second SIGHAN workshop on Chinese language processing. Association for Computational Linguistics, vol. 17 2003.
[11] L. Shala, V. Rus, and A. Graesser, "Automatic Speech Act Classification In Arabic," In Proceeding of Subjetividad y Procesos Cognitivos Conference 2010, pp. 284-292.
[12] Y. Bahou, L. H. Belguith, and A. B. Hamadou, "Towards a Human-Machine Spoken Dialogue in Arabic," In Proceeding of Workshop on HLT & NLP within the Arabic world: Arabic Language and local languages processing: Status Updates and Prospects, at the 6th Language Resources and Evaluation Conference (LREC'08), Marrakech, Maroc, 2008.
[13] C. Lhioui, A. Zouaghi, and M. Zrigui, "A Combined Method Based on Stochastic and Linguistic Paradigm for the Understanding of Arabic Spontaneous Utterances," In Proceeding of CICLing 2013, Computational Linguistics and Intelligent Text Processing Lecture Notes in Computer Science, Samos, Greece, 2013, pp. 549-558.
[14] M. Hijjawi, Z. Bandar, and K. Crockett, "User's Utterance Classification Using Machine Learning for Arabic Conversational Agents," In Proceeding of 5th International Conference on Computer Science and Information Technology (CSIT), 2013, pp. 223-232.
[15] M. Hijjawi, Z. Bandar, K. Crockett, and D. Mclean, "ArabChat: an Arabic Conversational Agent," In Proceeding of 6th International Conference on Computer Science and Information Technology (CSIT), 2014, pp. 227-237.
[16] L. Belguith, L. Baccour, and G. Mourad, "Segmentation de textes arabes basée sur l'analyse contextuelle des signes de ponctuations et de certaines particules," In Proceeding of 12th Conference on Natural Language Processing (TALN'2005), 2005, pp. 451-456.
[17] A. A. Touir, H. Mathkour, and W. Al-Sanea, "Semantic-Based Segmentation of Arabic Texts. ," Information Technology Journal, vol. 7 (7), 2008.

[18] I. Keskes, F. Benamara, and L. H. Belguith, "Clause-based Discourse Segmentation of Arabic Texts," In Proceeding of The eighth international conference on Language Resources and Evaluation (LREC), Istanbul, 2012.

[19] I. Khalifa, Z. A. Feki, and A. Farawila, "Arabic Discourse Segmentation Based on Rhetorical Methods," International Journal of Electric & Computer Sciences IJECS-IJENS, vol. 11 (01), 2011.

[20] A. A. Elmadany, S. M. Abdou, and M. Gheith, "Recent Approaches to Arabic Dialogue Acts Classifications," 4th International Conferences on Natural Language Processing (NLP-2015) - Computer Science & Information Technology (CS & IT) Series, vol. 5 (4), pp. 117–129, 2015.

[21] A. A. Elmadany, S. M. Abdou, and M. Gheith, "Arabic Inquiry-Answer Dialogue Acts Annotation Schema," IOSR Journal of Engineering (IOSRJEN), vol. 04 (12-V2), pp. 32-36, 2014.

[22] T. Kudo and Y. Matsumoto, "Chunking with support vector machines," In Proceeding of NAACL-01, 2001.

[23] Y. Kudo and Y. Matsumoto, "Use of Support Vector Learning for Chunk Identification " In Proceeding of CoNLL, Lisbon, Portugal, 2000, pp. 142-144.

[24] M. A. Meselhi, H. M. A. Bakr, I. Ziedan, and K. Shaalan, "A Novel Hybrid Approach to Arabic Named Entity Recognition," presented at the The 10th China Workshop on Machine Translation (CWMT 2014), 2014.

[25] S. Pradhan, K. Hacioglu, V. Krugler, W. Ward, J. H. Martin, and D. Jurafsky, "Support vector learning for semantic argument classification," Machine Learning 60, vol. Machine Learning (1-3), pp. 11-39, 2005.